\pdfoutput=1

\documentclass[11pt]{article}

\usepackage[final]{acl}

\usepackage{times}
\usepackage{latexsym}
\usepackage[T1]{fontenc}
\usepackage[utf8]{inputenc}
\usepackage{microtype}
\usepackage{inconsolata}
\usepackage{graphicx}
\usepackage{amsmath}
\usepackage{amssymb}
\usepackage{amsfonts}
\usepackage{booktabs}
\usepackage{multirow}
\usepackage{enumitem}
\usepackage{pifont}
\usepackage{xcolor}

\DeclareMathOperator{\activation}{activation}

\DeclareMathOperator{\sigmoid}{sigmoid}
\DeclareMathOperator{\layernorm}{ln}
\DeclareMathOperator{\dropout}{dropout}

\newcommand{\vect}[1]{\mathbf{#1}}
\newcommand{\mat}[1]{\mathbf{#1}}
\newcommand{\head}[1]{\paragraph{#1}}
\newcommand{\specialtoken}[1]{\texttt{#1}}
\newcommand{\hfmodel}[1]{#1}
\newcommand{\library}[1]{\texttt{#1}}
\newcommand{\entity}[1]{\textit{#1}}
\newcommand{\itemseprule}[0]{\vspace{0.3em}\hrule\vspace{0.3em}}
\newcommand{\hlink}[1]{\textit{\underline{#1}}}
\newcommand{\cmark}{{\textcolor{green!70!black}{\ding{52}}}}
\newcommand{\xmark}{{\textcolor{red!80!black}{\ding{54}}}}
\newcommand{\tableindent}{~~}
\newcommand{\scorewithinterval}[2]{\begin{tabular}{@{}r@{}}#1\\[-1mm]\small{$\pm$#2}\end{tabular}}

\title{Dynamic Injection of Entity Knowledge into Dense Retrievers}

\author{
Ikuya Yamada${}^{1,2}$\hspace{1.5mm}
Ryokan Ri$^{3}$\hspace{1.5mm}
Takeshi Kojima$^4$\hspace{1.5mm}
Yusuke Iwasawa$^4$\hspace{1.5mm}
Yutaka Matsuo$^4$
\\[2.5mm]
$^1$Studio Ousia\quad
$^2$RIKEN AIP\quad
$^3$SB Intuitions\quad
$^4$The University of Tokyo
\\[2.5mm]
\texttt{ikuya@ousia.jp} \quad 
\texttt{ryou0634@gmail.com} \\
\texttt{\{t.kojima,iwasawa,matsuo\}@weblab.t.u-tokyo.ac.jp} \\
}
\begin{document}
\maketitle
\begin{abstract}
Dense retrievers often struggle with queries involving less-frequent entities due to their limited entity knowledge.  
We propose the Knowledgeable Passage Retriever (KPR), a BERT-based retriever enhanced with a context-entity attention layer and dynamically updatable entity embeddings.  
This design enables KPR to incorporate external entity knowledge without retraining.  
Experiments on three datasets demonstrate that KPR consistently improves retrieval accuracy, with particularly large gains on the EntityQuestions dataset.
When built on the off-the-shelf \hfmodel{bge-base} retriever, KPR achieves state-of-the-art performance among similarly sized models on two datasets.  
Models and code are released at \href{https://github.com/knowledgeable-embedding/knowledgeable-embedding}{\nolinkurl{github.com/knowledgeable-embedding/knowledgeable-embedding}}.
\end{abstract}

\section{Introduction}

Language models (LMs) struggle to capture less-frequent or up-to-date entity knowledge \cite{pmlr-v202-kandpal23a,mallen-etal-2023-trust}, often resulting in hallucinations \cite{shuster-etal-2021-retrieval-augmentation}.
Retrieval-augmented generation (RAG), which enhances LMs by leveraging external knowledge retrieval, is a promising approach to mitigate this issue.
Dense retrievers are commonly employed for this purpose; however, because they also rely on LMs, they likewise struggle with queries involving less-frequent entities \cite{sciavolino-etal-2021-simple} and often fail to retrieve such knowledge effectively.

\begin{figure}[t]
  \centering
  \includegraphics[width=\linewidth]{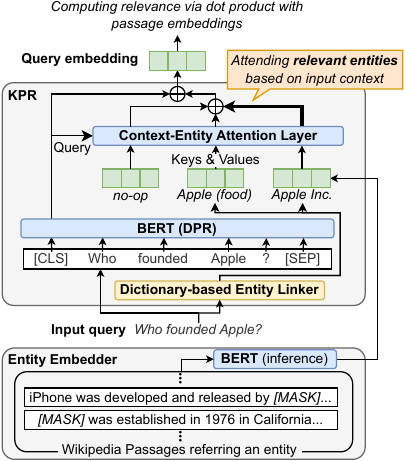}
  \caption{Architecture of KPR, a dense retriever with a context-entity attention layer that incorporates entity knowledge, given a query as input.
Entity embeddings are obtained via BERT inference, and both the embeddings and the entity linker can be updated without retraining.
Relevance is scored by the dot product of query and passage embeddings, both produced by KPR.
}
  \label{fig:architecture}
\end{figure}

In this paper, we address this problem by proposing a simple extension to dense retrievers that dynamically injects entity knowledge into their embeddings.  
Specifically, we introduce the \underline{K}nowledgeable \underline{P}assage \underline{R}etriever (KPR), a BERT-based dense retriever \cite{karpukhin-etal-2020-dense} that integrates entity embeddings adaptable at inference time (see Figure~\ref{fig:architecture}).  
KPR is intentionally designed with a simple architecture and is trained to attend to entity knowledge based on the context of the input text, through a context-entity attention layer placed on top of BERT. 
Entity embeddings are obtained via single-pass BERT inference on texts referring to the entity and are kept frozen during training.  
KPR detects entities in the input text using a simple dictionary-based entity linker.  
Since both the entity embeddings and the linker can be modified after training, new entity knowledge can be easily added or updated without retraining.

To evaluate KPR, we use the EntityQuestions (EQ) dataset \cite{sciavolino-etal-2021-simple}, which includes many queries with less-frequent entities, as well as the Natural Questions (NQ) \cite{kwiatkowski-etal-2019-natural} and TriviaQA (TQA) \cite{joshi-etal-2017-triviaqa} datasets.  
KPR achieves a substantial 12.6\% gain in top 20 accuracy on EQ over the model without KPR extensions and consistently improves performance on the other two datasets.  
We further build KPR on top of the off-the-shelf \hfmodel{bge-base} retriever, achieving state-of-the-art results on EQ and TQA compared to other retrievers of similar size.

\section{Method}
\label{subsec:method}

KPR is built upon DPR \cite{karpukhin-etal-2020-dense}, a widely used BERT-based dense retriever.  
We add an attention layer on top of DPR to incorporate entity knowledge from a knowledge base (KB) (see Figure~\ref{fig:architecture}).  
We use Wikipedia as the target KB.

\head{DPR.} 
DPR encodes a query or passage into a $D$-dimensional embedding, obtained from the output embedding of BERT’s \specialtoken{[CLS]} token.  
Relevance between a query and a passage is scored as the dot product of their embeddings.  
The model is fine-tuned on datasets containing queries paired with positive and negative passages, using a cross-entropy loss with the dot product scores as logits.  
See Appendix~\ref{subsec:dpr} for further details.

\head{Input entity representation.} KPR uses the following two \( D \)-dimensional input embeddings:
\begin{itemize}[leftmargin=1.5em]
\item \textit{Entity embedding}, assigned to each entity in the KB, represents the entity itself.

\item \textit{Entity position embedding}, assigned to each position in the input tokens, encodes positional information.
\end{itemize}
KPR takes the input tokens and a set of entities \( E = e_1, \ldots, e_N \) detected from the text using an entity linker.  
The input representation of each entity is computed by summing its entity embedding and entity position embedding based on its position in the input token sequence.  
If an entity spans multiple tokens, its position embedding is computed as the average of the embeddings of the corresponding positions \cite{yamada-etal-2020-luke}.  
To prevent the entity sequence from being empty, a \( D \)-dimensional \textit{no-op} embedding is appended to the sequence of input entity representations.

\head{Context-entity attention layer.} 
KPR adopts a single-head key-query-value attention mechanism \cite{NIPS2017_3f5ee243}.
The query matrix \( \mat{Q} \) is computed based on the output embedding of BERT's \specialtoken{[CLS]} token (denoted by \( \mat{H}_{\text{\specialtoken{[CLS]}}} \in \mathbb{R}^{1 \times D} \)).  
The key and value matrices, \( \mat{K} \) and \( \mat{V} \), are computed based on the input entity representations denoted by \( \mat{U} \in \mathbb{R}^{N+1 \times D} \):
\begin{equation*}
  \mat{Q} = \mat{H}_{\text{\specialtoken{[CLS]}}}\mat{X}_q,\,\,\,\mat{K} = \mat{U}\mat{X}_k,\,\,\,\mat{V} = \mat{U}\mat{X}_v,
\end{equation*}
where $\mat{X}_q \in \mathbb{R}^{D \times D}$, $\mat{X}_k \in \mathbb{R}^{D \times D}$, and $\mat{X}_v \in \mathbb{R}^{D \times D}$ are weight matrices.
We aim for the attention mechanism to attend to useful entities based on the context by using entity embeddings as the key and the embedding of the \specialtoken{[CLS]} token, which encodes the context of the input text, as the query.  

KPR computes the output embedding \( \mat{Z} \in \mathbb{R}^{1 \times D} \) as:
\begin{align*}
    \mat{Y} &= \activation\left(\frac{\mat{Q} \mat{K}^\top}{\sqrt{D}}\right) \mat{V}, \\
    \mat{Z} &= \layernorm\left(\dropout\left(\mat{Y}\right) + \mat{H}_{\text{\specialtoken{[CLS]}}}\right),
\end{align*}
where \( \activation(\cdot) \), \( \layernorm(\cdot) \), and \( \dropout(\cdot) \) denote the activation, layer normalization, and dropout functions, respectively.  
We use the sigmoid function with a length-based bias, \( \sigmoid(x - \log N + 1) \) \cite{ramapuram2025theory}, as the activation. 

\head{Entity embedder.}
We compute entity embeddings based on BERT inference over Wikipedia passages that refer to the corresponding entity \cite{ye-etal-2022-simple}.  
Specifically, for each passage containing an anchor link to the entity, we replace the entity name with BERT’s \specialtoken{[MASK]} token, extract the output embedding of the \specialtoken{[MASK]} token, and average the resulting embeddings across passages.  
The final embeddings are then normalized to match the average norm of BERT’s input token embeddings.

\head{Entity linker.}
We use dictionary-based string matching to detect entities mentioned in the text.  
The dictionary comprises anchor names and their possible referent entities, obtained from Wikipedia hyperlinks.  
For ambiguous names (e.g., “Apple,” which can refer to \entity{Apple Inc.} and \entity{Apple (food)}), we do not disambiguate to a single entity, but instead input all possible candidates to maintain recall.

\head{Training.} 
The parameters in our attention layer, entity position embeddings, and no-op embedding are initialized randomly.  
Entity embeddings are kept frozen during training, enabling them to be added or updated dynamically after training.  
The dropout probability in the attention layer is set to match that of the other layers.  
We follow DPR \cite{karpukhin-etal-2020-dense} for the remaining training settings including the loss function.
Note that entities can be dynamically updated in KPR without retraining, since KPR relies on entity embeddings computed via BERT inference, which are kept frozen during training, and on a dictionary-based entity linker, whose entries can be modified.

\head{Computational overhead.}
KPR adds entity embeddings to BERT, requiring additional storage.  
However, since the embeddings are used only as input features, they can be implemented as a sparse lookup table and stored not only in GPU memory but also in CPU memory or even on disk, with minimal impact on speed~\cite{yu2025scaling}.  
Furthermore, KPR introduces only a small overhead in total FLOPs, as discussed in Appendix~\ref{subsec:computational-overhead}. 

\section{Experiments}
\label{sec:experiments}

\head{Entity set.}
We use 7.2M English Wikipedia entities to construct our entity embeddings and linker.

\head{Datasets and metrics.} We train our model on the dataset proposed by \citet{karpukhin-etal-2020-dense}, which is based on NQ, TQA, WebQuestions~\cite{berant-etal-2013-semantic}, and CuratedTREC~\cite{10.1007/978-3-319-24027-5_20}.  
We evaluate the models on the EQ, NQ, and TQA datasets.  
Following \citet{sciavolino-etal-2021-simple}, we report top 20 retrieval accuracy.
Top 100 accuracies are also provided in Appendix \ref{subsec:full-results}.
We use the 21M Wikipedia passages released by \citet{karpukhin-etal-2020-dense} as the target passages.  

\head{Baselines.}  
We use vanilla DPR and conventional BM25, which performs robustly on queries with less frequent entities~\cite{sciavolino-etal-2021-simple}.

\head{Base models.} We use base-sized BERT and RetroMAE \cite{xiao-etal-2022-retromae}, a state-of-the-art BERT-based model pretrained for retrieval.  
For DPR, we also adopt PELT \cite{ye-etal-2022-simple}, a BERT-based model that injects entity knowledge by inserting each entity embedding, placed between the token embeddings of parentheses, immediately after the corresponding entity token.  
While PELT introduces entity knowledge through input augmentation, KPR incorporates it via a dedicated attention layer.  
Unlike KPR, PELT requires each input entity name to be disambiguated to a single entity, which incurs additional computational complexity.

\head{Entity embeddings.} We use the same entity embeddings, derived from base-sized BERT inference (\S\ref{subsec:method}), for both PELT and all KPR models.

\head{Entity linker.} By default, we use our dictionary-based linker (§\ref{subsec:method}) and additionally employ the state-of-the-art ReFinED linker \cite{ayoola-etal-2022-refined}.  
We select ReFinED over other off-the-shelf systems such as BLINK \cite{wu-etal-2020-scalable}, GENRE \cite{cao2021autoregressive},  and LUKE \cite{yamada-etal-2022-global} due to its superior performance.

We refer to KPR based on BERT as KPR$_\text{BERT}$ and name the other models accordingly. 
Further details are provided in Appendix~\ref{subsec:detailed-experimental-setup}.

\subsection{Results and Analysis}
\label{subsec:results}

\begin{table}[t]
\centering
\scalebox{0.67}{
\begin{tabular}{l|c|c|cccc}
\toprule
    \textbf{Model} & \textbf{\shortstack{\textbf{Base} \\ \textbf{Model}}} & \textbf{\shortstack{\textbf{Entity} \\ \textbf{Linker}}} & \textbf{EQ} & \textbf{NQ} & \textbf{TQA} & \textbf{Avg} \\ 
    \midrule
    BM25 & -- & -- & 71.2 & 59.1 & 66.9 & 66.0 \\
    \midrule
    DPR$_\text{BERT}$ & BERT & -- & \scorewithinterval{56.8}{0.3} & \scorewithinterval{79.3}{0.2} & \scorewithinterval{78.9}{0.1} & 71.7 \\
    KPR$_\text{BERT}$ & BERT & Dictionary & \scorewithinterval{\underline{69.4}}{0.4} & \scorewithinterval{\underline{80.7}}{0.3} & \scorewithinterval{\underline{79.7}}{0.1} & \underline{76.6} \\
    \midrule
    DPR$_\text{PELT}$ & BERT & ReFinED & \scorewithinterval{63.8}{0.8} & \scorewithinterval{79.8}{0.2} & \scorewithinterval{79.0}{0.1} & 74.2 \\
    KPR$_\text{BERT}$ & BERT & ReFinED & \scorewithinterval{\underline{69.0}}{0.4} & \scorewithinterval{\underline{80.4}}{0.2} & \scorewithinterval{\underline{79.8}}{0.1} & \underline{76.4} \\
    \midrule
    DPR$_\text{RetroMAE}$ & RetroMAE & -- &  \scorewithinterval{70.9}{0.2} & \scorewithinterval{81.0}{0.2} & \scorewithinterval{81.2}{0.1} & 77.7 \\
    KPR$_\text{RetroMAE}$ & RetroMAE & Dictionary & \scorewithinterval{\textbf{\underline{74.7}}}{0.4} & \scorewithinterval{\textbf{\underline{83.0}}}{0.3} & \scorewithinterval{\textbf{\underline{82.1}}}{0.1} & \textbf{\underline{79.9}} \\
\bottomrule
\end{tabular}
}
\caption{
Top 20 accuracies of KPR and baseline models. For clarity, models are grouped by their base pretrained models and entity linkers. We report mean accuracy and 95\% confidence intervals based on Student's t-distribution over 5 training runs with different random seeds.
The best overall mean scores are in bold; the best within each group are underlined. Top 100 accuracies are provided in Table~\ref{table:results_main_full}.}
\label{table:results_main}
\end{table}

\begin{table}[t]
\centering
\scalebox{0.67}{
\begin{tabular}{l|cccc}
\toprule
    \textbf{Model} & \textbf{EQ} & \textbf{NQ} & \textbf{TQA} & \textbf{Avg} \\ 
    \midrule
    \textbf{Entity embeddings:} \\
    \tableindent\tableindent Random initialization & 64.9 & 80.3 & 79.0 & 74.7 \\
    \tableindent\tableindent Wikipedia2Vec & 65.9 & 80.4 & 79.2 & 75.2 \\
    \tableindent\tableindent BERT intermediate layer \#3 & 66.2 & 80.3 & 79.4 & 75.3 \\
    \tableindent\tableindent BERT intermediate layer \#6 & 66.8 & 80.3 & 79.4 & 75.5 \\
    \tableindent\tableindent BERT intermediate layer \#9 & 67.8 & 80.4 & 79.6 & 75.9\\
    \tableindent\tableindent BERT last layer (KPR$_\text{BERT}$) & \underline{69.3} & \underline{80.6} & \underline{79.7} & \underline{76.5} \\
    \midrule
    \textbf{Activation functions:} \\
    \tableindent\tableindent Softmax function & 66.3 & 80.5 & 79.5 & 75.4\\
    \tableindent\tableindent Sigmoid function with length bias & \underline{69.3} & \underline{80.6} & \underline{79.7} & \underline{76.5} \\
    \bottomrule
\end{tabular}
}
\caption{Top 20 accuracy of KPR$_\text{BERT}$ with different entity embeddings and activation functions.  
Unlike Table~\ref{table:results_main}, the results are based on a single training run.  
The best score in each group is underlined.}
\label{table:analysis}
\end{table}

\begin{figure}[t]
  \centering
  \includegraphics[width=\linewidth]{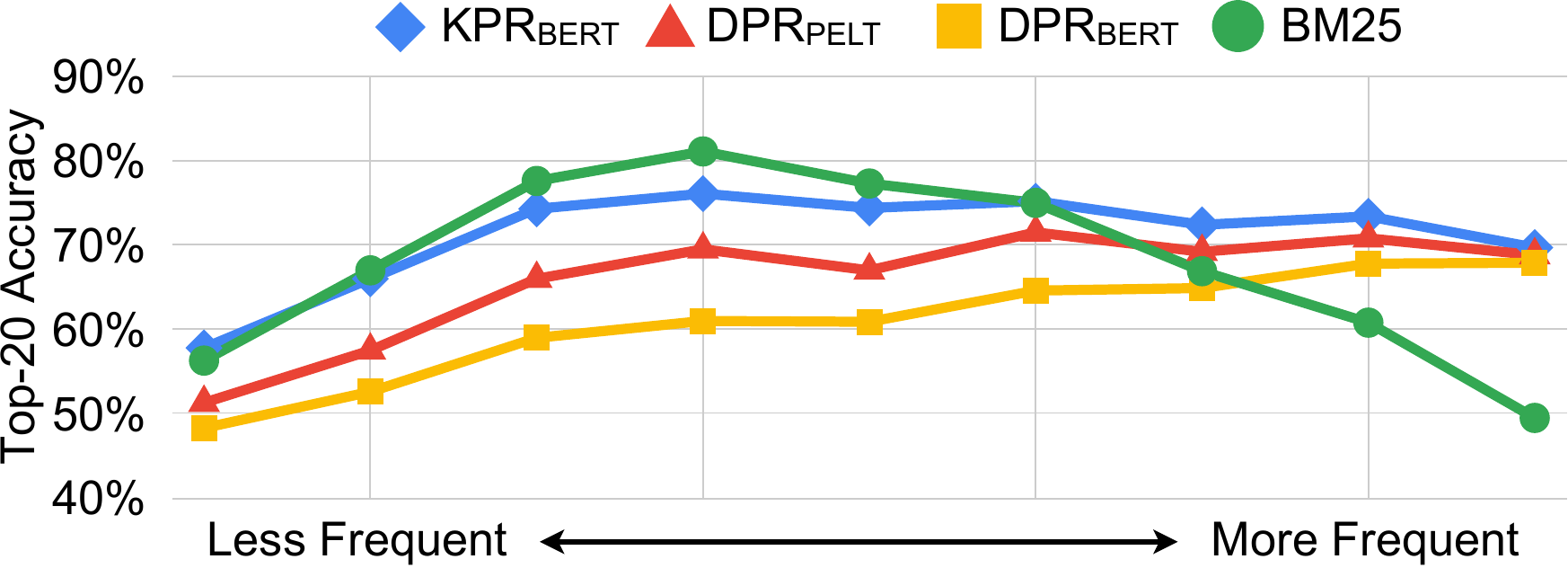}
  \caption{Top 20 retrieval accuracy of KPR$_\text{BERT}$, DPR$_\text{PELT}$, DPR$_\text{BERT}$, and BM25 on EQ, grouped into 10 bins based on entity frequency in Wikipedia.}
  \label{fig:eq-analysis}
\end{figure}

\begin{table}[t]
\centering
\scalebox{0.67}{
\begin{tabular}{l|cccc}
\toprule
    \textbf{Model} & \textbf{EQ} & \textbf{NQ} & \textbf{TQA} & \textbf{Avg} \\ 
    \midrule
    contriever \cite{izacard2022unsupervised} & 63.0 & 67.9 & 73.9 & 68.3 \\
    gte-base \cite{li2023towards} & 70.8 & 77.0 & 76.6 & 74.7\\
    e5-base \cite{wang2022text} & 72.4 & \textbf{86.2} & 81.4 & 80.0 \\
    bge-base \cite{10.1145/3626772.3657878} & 71.0 & 82.3 & 80.0 & 77.8  \\
    KPR$_\text{bge-base}$ & \textbf{76.8} & 82.4 & \textbf{81.5} & \textbf{80.2} \\
\bottomrule
\end{tabular}
}
\caption{
Top 20 accuracies of KPR based on \hfmodel{bge-base} and off-the-shelf retrievers on EQ, NQ, and TQA. Top 100 accuracies are provided in Table~\ref{table:results_sota_full}.}
\label{table:results_sota}
\end{table}

\head{Main results.} 
Table~\ref{table:results_main} shows that KPR significantly outperforms all baselines across datasets and base models.  
KPR$_\text{BERT}$ surpasses DPR$_\text{BERT}$ by 12.6\% on EQ and 4.9\% on average.  
Using the same ReFinED linker, KPR$_\text{BERT}$ consistently outperforms DPR$_\text{PELT}$, with a 5.2\% gain on EQ, suggesting that the entity knowledge injected by PELT is partially lost within BERT and highlighting the benefit of KPR's attention layer placed on top.  
KPR$_\text{RetroMAE}$ achieves the best performance across all datasets, demonstrating its effectiveness even with a strong pretrained model.  
The gain is notably larger on EQ in all settings, as it includes many queries involving less-frequent entities.  
Further analysis of the EQ results is provided below.

Additional results on the MS MARCO dataset are presented in Appendix~\ref{subsec:ms-marco}.

\head{Effects of entity linker.} Table~\ref{table:results_main} also shows that KPR$_\text{BERT}$ with the dictionary-based linker outperforms its ReFinED-based variant on average.  
This is somewhat surprising, as the dictionary linker simply applies string matching without disambiguation (§\ref{subsec:method}) and may detect incorrect or noisy entities.  
We attribute this to ReFinED’s slightly lower recall, as it extracts entities only when confident.  
For example, ReFinED detects 0.93 entities per query on average in EQ, compared to 0.97 with the dictionary linker.  
These results also suggest that KPR is robust to noise, likely due to its attention mechanism’s ability to focus on contextually relevant entities.  
See Appendix~\ref{subsec:analysis-attention} for further analysis.

\head{How do entity embeddings affect performance?}  
We evaluate two baselines for computing entity embeddings: random initialization and Wikipedia2Vec embeddings \cite{yamada-etal-2020-wikipedia2vec}, both based on the same Wikipedia dump as our entity embeddings.  
We also test embeddings extracted from an intermediate BERT layer, motivated by prior work suggesting that entity knowledge is well captured there \cite{NEURIPS2022_6f1d43d5}.

Table~\ref{table:analysis} shows that BERT-based embeddings consistently outperform the alternatives.  
Notably, using the last layer yields the best performance across all datasets.  
This is likely because \( \mat{H}_{\text{\specialtoken{[CLS]}}} \), used in our context-entity attention layer, is also taken from the last layer, which may help the model better capture relevance between context and entities.

\head{Sigmoid vs. softmax.}  
We evaluate the effect of replacing the sigmoid activation in our attention layer with the softmax function, which is commonly used in attention mechanisms.  
Table~\ref{table:analysis} shows that sigmoid consistently outperforms softmax across all datasets.  
We attribute this to the fact that sigmoid allows the model to assess each entity’s relevance independently, without being influenced by the presence of other relevant entities.

\head{Analysis of EQ results.}  
To examine whether KPR's improvement stems from incorporating knowledge of less-frequent entities, we divide the EQ examples into 10 bins based on the frequency of the entity in each query and measure performance within each bin.  
This is feasible because every EQ query contains a single entity.  
We obtain entity frequencies from Wikipedia hyperlinks and create 10 log-spaced bins ranging from 1 to 10{,}000.

Figure~\ref{fig:eq-analysis} shows that KPR$_\text{BERT}$ is more consistent across entity frequencies than other models and outperforms DPR$_\text{BERT}$ and DPR$_\text{PELT}$, especially on queries with less-frequent entities.  
Compared to BM25, it performs comparably on less-frequent entities and substantially better on frequent ones.

\head{Analysis of KPR's Attention Mechanism.}  
Appendix~\ref{subsec:analysis-attention} provides a qualitative analysis of our attention mechanism.  
We observe that KPR tends to assign lower weights to common entities and generally higher weights to correct entities than to incorrect ones, as shown in Figure~\ref{fig:attention-analysis}.

\subsection{Pushing State-of-the-Art}
\label{subsec:pushing-sota}

To evaluate the effectiveness of KPR on recent off-the-shelf retrievers, we select the \hfmodel{bge-base} model \cite{10.1145/3626772.3657878} and train KPR$_\text{bge-base}$ using it as the base model.  
Since \hfmodel{bge-base} is already trained on large-scale, high-quality datasets, we freeze the base model to prevent catastrophic forgetting and train only the newly introduced parameters.  
Detailed settings are provided in Appendix~\ref{subsec:detailed-experimental-setup}.

Table~\ref{table:results_sota} presents results comparing our model with recent off-the-shelf retrievers.  
KPR$_\text{bge-base}$ consistently outperforms \hfmodel{bge-base} across all datasets, with a substantial 5.8\% gain on EQ.  
It also achieves the highest average performance among all strong off-the-shelf retrievers.

\section{Related Work}
The performance of dense retrievers \cite{karpukhin-etal-2020-dense,pmlr-v119-guu20a} remains insufficient for queries involving less-frequent entities \cite{sciavolino-etal-2021-simple}, due to the limited knowledge of such entities in LMs \cite{pmlr-v202-kandpal23a,mallen-etal-2023-trust}.
Several studies have explored incorporating entity knowledge to enhance LMs \cite{zhang-etal-2019-ernie,yamada-etal-2020-luke,ye-etal-2022-simple,zhang-etal-2023-plug}, among which we adopt PELT \cite{ye-etal-2022-simple} as a baseline.

Entity knowledge has also been leveraged to improve retrieval tasks \cite{10.1007/s10791-015-9267-x,10.1145/3077136.3080768,10.1145/3038912.3052558,liu-etal-2018-entity,10.1145/3511808.3557285,nguyen-etal-2024-dyvo,10.1007/978-3-031-56027-9_13}.  
However, to our knowledge, only \citet{10.1145/3511808.3557285} is directly applicable to dense retrieval, integrating Wikipedia2Vec embeddings into a BERT-based retriever.  
Because its entity representations are computed independently of BERT, the model cannot learn to select relevant entities based on context.  
As a result, it fails to improve performance compared to the vanilla model when passages are represented by a single embedding.  
Moreover, its effectiveness on queries involving less-frequent entities remains unexamined, and updating entity embeddings requires retraining Wikipedia2Vec.  
In contrast, KPR integrates an attention layer into BERT, enabling it to select contextually relevant entities more effectively.  
This leads to improved performance across multiple retrieval benchmarks.

\section{Conclusions}

We proposed KPR, a dense retriever with a context-entity attention layer and dynamically updatable entity knowledge.  
KPR consistently improves performance across benchmarks, particularly on queries involving less-frequent entities.  
When built on \hfmodel{bge-base}, it achieves state-of-the-art results on two benchmarks among similarly sized models.  
Future work includes applying KPR to decoder-based retrievers and extending it to KBs beyond Wikipedia.

\section*{Limitations}

This work focuses on English-language datasets and assumes the availability of a KB, namely Wikipedia.
While the proposed method is modular by design, its effectiveness in languages other than English or in domains lacking a comprehensive KB remains unexplored.

\bibliography{references}

\clearpage
\appendix

\section{Overview of DPR}
\label{subsec:dpr}

DPR is a BERT-based model that encodes a query $q$ and a passage $p$ into $D$-dimensional embeddings, denoted as $\vect{e}_q \in \mathbb{R}^D$ and $\vect{e}_p \in \mathbb{R}^D$.  
The embedding is taken from the output embedding of BERT’s \specialtoken{[CLS]} token.  
To construct passage $p$, the passage title and text are concatenated with BERT's \specialtoken{[SEP]} tokens as: \specialtoken{[CLS]} \textit{passage title} \specialtoken{[SEP]} \textit{passage text} \specialtoken{[SEP]}.  
Given query $q$, the relevance of passage $p$ is computed as the dot product $\langle \vect{e}_q, \vect{e}_p \rangle$.

\head{Training.}
Let $\mathcal{D} = \{ \langle q_i, p^+_i, p^-_{i,1}, \cdots, p^-_{i,N} \rangle \}_{i=1}^M$ be a set of $M$ training instances, where each instance consists of a query $q_i$, a positive passage $p^+_i$ relevant to the query, and $N$ negative passages $p^-_{i,j}$ irrelevant to the query.  
The model is trained by minimizing the negative log-likelihood of the positive passage:
\begin{equation*}
  \resizebox{1.0\hsize}{!}{
  $\mathcal{L} = -\log \frac{\exp(\langle \mathbf{e}_{q_i},\, \mathbf{e}_{p_i^+} \rangle)}{\exp(\langle \mathbf{e}_{q_i},\, \mathbf{e}_{p_i^+} \rangle) + \sum_{j=1}^N \exp(\langle \mathbf{e}_{q_i},\, \mathbf{e}_{p^-_{i,j}} \rangle)}$.
  }
  \label{eq:nll-loss}
\end{equation*}
During training, \textit{in-batch negatives} are used, where each positive and negative passage in the batch serves as a negative for all other queries.
 
\head{Inference.}
DPR constructs a passage index by encoding all target passages.  
At runtime, it retrieves the top-ranked passages using maximum inner product search, with the query embedding as input.

\section{Notes on Computational Overhead}
\label{subsec:computational-overhead}

In this section, we analyze the impact of KPR’s attention layer on the total number of floating point operations (FLOPs), following the estimation method of \citet{kaplan2020scaling}.  
The non-embedding FLOPs for a single forward pass of BERT can be approximated as:
\begin{equation*}
    \text{FLOPs}_\text{BERT} \approx 2LDM(12D + M),
\end{equation*}
where $L$, $D$, and $M$ denote the number of hidden layers, hidden size, and input token length, respectively.

The FLOPs for a forward pass of KPR’s attention layer can be approximated as:
\begin{equation*}
    \text{FLOPs}_\text{KPR-att} \approx 2D^2(2N+1) + 2DN,
\end{equation*}
where $N$ denotes the number of input entities. The two terms correspond to the computation of the key-query-value matrices and the dot product, respectively.

The total FLOPs for a single forward pass of KPR is then given by:
\begin{equation*}
    \text{FLOPs}_\text{KPR} \approx \text{FLOPs}_\text{BERT} + \text{FLOPs}_\text{KPR-att}.
\end{equation*}

For example, with $M = 128$ input tokens and $N = 16$ entities, a forward pass of base-sized BERT ($L = 12$, $D = 768$) requires approximately 22 GFLOPs.  
The additional cost from KPR’s attention layer is about 39 MFLOPs, accounting for only 0.18\% of the total FLOPs.

Note that while FLOPs provide a hardware-agnostic estimate of computational cost, actual runtime may vary due to factors such as implementation optimizations and hardware constraints. Consequently, the computational latency of KPR’s attention layer may not precisely correspond to its 0.18\% FLOPs share.

\begin{table*}[t]
\centering
\scalebox{0.9}{
\begin{tabular}{l|cccc}
\toprule
    \textbf{Dataset} & \textbf{License} & \textbf{\#Train} & \textbf{\#Dev} & \textbf{\#Test} \\ 
    \midrule
    Natural Questions & Apache-2.0 & 58,880 & 8,757 & 3,610 \\
    TriviaQA & Apache-2.0 & 60,413 & -- & 11,313 \\
    WebQuestions & CC BY 4.0 & 2,474 & -- & -- \\
    CuratedTREC & -- & 1,125 & -- & -- \\
    EntityQuestions & MIT & -- & -- & 22,075 \\
    MS MARCO & Non-commercial & 502,939 & 6,980 & -- \\
    GraphQuestions & CC BY 4.0 & 1,672 & 417 & 2,075 \\
\bottomrule
\end{tabular}
}
\caption{
Licenses and the number of training, development, and test examples (if available and used) for each dataset used in this paper.}
\label{table:datasets}
\end{table*}

\begin{table*}[t]
\centering
\scalebox{0.9}{
\begin{tabular}{l|c|c|cccc}
\toprule
    \textbf{Model} & \textbf{Base Model} & \textbf{Entity Linker} & \textbf{EQ} & \textbf{NQ} & \textbf{TQA} & \textbf{Avg} \\
    \midrule
    BM25 & -- & -- & 79.8 & 73.7 & 76.7 & 76.7 \\
    \midrule
    DPR$_\text{BERT}$ & BERT & -- & 70.0 & 86.0 & 84.7 & 80.2 \\
    KPR$_\text{BERT}$ & BERT & Dictionary & \underline{78.9} & \underline{87.0} & \underline{85.0} & \underline{83.6} \\
    \midrule
    DPR$_\text{PELT}$ & BERT & ReFinED & 75.3 & 86.3 & 84.7 & 82.1 \\
    KPR$_\text{BERT}$ & BERT & ReFinED & \underline{78.7} & \underline{86.7} & \underline{85.0} & \underline{83.5} \\
    \midrule
    DPR$_\text{RetroMAE}$ & RetroMAE & -- & 79.8 & 87.7 & 86.0 & 84.5 \\
    KPR$_\text{RetroMAE}$ & RetroMAE & Dictionary & \textbf{\underline{82.3}} & \textbf{\underline{88.2}} & \textbf{\underline{86.7}} & \textbf{\underline{85.7}} \\
\bottomrule
\end{tabular}
}
\caption{
Top 100 accuracies of KPR and baseline models. For clarity, models are grouped by their base pretrained models and entity linkers. The best overall scores are shown in bold, and the best scores within each group are underlined. The corresponding top 20 accuracies are available in Table~\ref{table:results_main}.}
\label{table:results_main_full}
\end{table*}

\begin{table*}[t]
\centering
\scalebox{0.9}{
\begin{tabular}{l|cccc}
\toprule
    \textbf{Model} & \textbf{EQ} & \textbf{NQ} & \textbf{TQA} & \textbf{Avg} \\ 
    \midrule
    contriever \cite{izacard2022unsupervised} & 75.2 & 80.6 & 82.9 & 80.0 \\
    gte-base \cite{li2023towards} & 79.8 & 86.4 & 83.9 & 83.4 \\
    e5-base \cite{wang2022text} & 82.5 & \textbf{90.8} & 86.2 & 86.5 \\
    bge-base \cite{10.1145/3626772.3657878} & 80.8 & 88.5 & 85.9 & 85.1 \\
    KPR$_\text{bge-base}$ & \textbf{84.2} & 89.0 & \textbf{86.6} & \textbf{86.6} \\
\bottomrule
\end{tabular}
}
\caption{
Top 100 accuracies of KPR based on \hfmodel{bge-base} and off-the-shelf retrievers on EQ, NQ, and TQA. The corresponding top 20 accuracies are available in Table~\ref{table:results_sota}.}
\label{table:results_sota_full}
\end{table*}

\begin{table*}[t]
\centering
\scalebox{0.9}{
\begin{tabular}{l|ccccc}
\toprule
    \textbf{Model} & \textbf{MRR@10} &
    \textbf{MRR@100} &
    \textbf{R@10} & 
    \textbf{R@100} & 
    \textbf{R@1000} \\ 
    \midrule
    DPR$_\text{BERT}$ \cite{xiao-etal-2022-retromae} & 31.7 & 32.8 & 58.0 & 85.7 & 96.0 \\
    KPR$_\text{BERT}$ & \textbf{33.1} & \textbf{34.2} & \textbf{60.1} & \textbf{86.5} & \textbf{96.1} \\
    \midrule
    DPR$_\text{RetroMAE}$ \cite{xiao-etal-2022-retromae} & 35.5 & 36.7 & 63.6 & 89.2 & 97.6 \\
    KPR$_\text{RetroMAE}$ & \textbf{37.3} & \textbf{38.4} & \textbf{66.4} & \textbf{91.2} & \textbf{98.3} \\
\bottomrule
\end{tabular}
}
\caption{
The experimental results of KPR and DPR on the MS MARCO dataset.}
\label{table:results_msmarco}
\end{table*}

\section{Detailed Experimental Setup}
\label{subsec:detailed-experimental-setup}

\head{Entity vocabulary.}
The entity vocabulary of KPR consists of 7.2M English Wikipedia entities that appear as hyperlinks at least once in the April 2024 dump.  
These hyperlinks are extracted using the \library{mwparserfromhell} library.\footnote{\url{https://github.com/earwig/mwparserfromhell}}

\head{Entity embeddings.}
As described in Section~\ref{subsec:method}, entity embeddings are obtained by running BERT inference on Wikipedia passages that refer to the corresponding entities.  
For each entity, we randomly select up to 128 such passages and compute the embedding using the method outlined in Section~\ref{subsec:method}.

For the Wikipedia2Vec entity embeddings used in Section~\ref{subsec:results}, we train the model with the default hyperparameters, except that the embedding dimension is set to 768 to match BERT’s input embeddings.  
The same Wikipedia dump is used as for our entity embeddings.

\head{Entity linker.}
The dictionary used in our entity linker is constructed directly from entity hyperlinks in Wikipedia.  
For example, if a hyperlink with the anchor text ``New York'' refers to the entity \entity{New York City}, we register ``New York'' as an entity name and \entity{New York City} as its possible referent.

To build the dictionary, we collect two statistics commonly used in the entity linking literature: \textit{link probability}, the probability that a name appears as a hyperlink in Wikipedia, and \textit{commonness}, the probability that a name refers to a specific entity \cite{10.1145/1321440.1321475,10.1145/1458082.1458150}.  
To filter out names unlikely to denote entities (e.g., function words), we exclude a name if its link probability is below 5\%.  
We also exclude an entity if its commonness with respect to the name is below 30\%.

We tokenize text using the default English tokenizer provided by the \library{SpaCy} library \cite{Honnibal_spaCy_Industrial-strength_Natural_2020} and extract entity names by matching all possible n-grams against the dictionary.  
Dictionary matching is implemented using a trie from the \library{marisa-trie} library\footnote{\url{https://github.com/pytries/marisa-trie}} and is performed in a case-insensitive manner.

In Section~\ref{sec:experiments}, we also use the ReFinED entity linker, employing the model trained on Wikipedia with default parameters.

\head{Model.} We use the base-sized BERT\footnote{\url{https://huggingface.co/google-bert/bert-base-uncased}} with 110M parameters. RetroMAE\footnote{\url{https://huggingface.co/Shitao/RetroMAE}} and \hfmodel{bge-base}\footnote{\url{https://huggingface.co/BAAI/bge-base-en}} are also based on this BERT model.  
KPR with the entity embeddings contains 5.6B parameters. 

\head{Datasets.}
The licenses and the number of training, development, and test examples for the datasets used in this paper are provided in Table~\ref{table:datasets}.  
All datasets are publicly and freely available for research purposes. Access to CuratedTREC is governed by the TREC organizers.  
MS MARCO is freely available for non-commercial research purposes.\footnote{\url{https://microsoft.github.io/msmarco/}}  
We also use Wikipedia, which is licensed under CC BY-SA 4.0.

For the target passages used in retrieval, we use a preprocessed version of English Wikipedia containing 21M passages, originally released by \citet{karpukhin-etal-2020-dense}.  
The corpus is based on the December 2018 Wikipedia dump, with semi-structured content such as tables, infoboxes, lists, and disambiguation pages removed.  
The remaining article text is segmented into non-overlapping chunks, each containing approximately 100 words.

\head{Training.}
Our training settings follow \citet{karpukhin-etal-2020-dense}.  
For each query, we use one positive and one hard negative passage, and construct a mini-batch of 128 queries.  
We share parameters between the query and passage encoders to reduce computational cost, as using separate encoders yields nearly identical results (within 0.2\% difference in top 20 accuracy across datasets) in our experiments, consistent with prior work~\cite{lee2023back,10.1145/3613447}.  
We optimize the model using Adam~\cite{kingma2014adam} with a learning rate of 2e-5 for 40 epochs.  
We apply in-batch negatives (see Appendix \ref{subsec:dpr}) during training.  
We do not perform hyperparameter tuning and instead use the same settings provided in the GitHub repository of \citet{karpukhin-etal-2020-dense}.\footnote{\url{https://github.com/facebookresearch/DPR/blob/main/conf/train/biencoder_nq.yaml}}

To construct KPR with the \hfmodel{bge-base} model presented in Section~\ref{subsec:pushing-sota}, we make slight modifications to align with the original settings of \hfmodel{bge-base}.  
Specifically, we use cosine similarity instead of the dot product for the similarity function and set the temperature of the softmax function to 0.02 before computing the cross-entropy loss during training.  
We also prepend the instruction ``Represent this sentence for searching relevant passages:'' to each query.  
We tune the hyperparameters over 12 trials using the development set of NQ.  
In particular, we select the number of epochs from $\{10, 20, 40\}$ and the learning rate from $\{2\mathrm{e}{-5}, 3\mathrm{e}{-5}, 5\mathrm{e}{-5}, 1\mathrm{e}{-4}\}$, and choose 20 epochs and a learning rate of $5\mathrm{e}{-5}$.  
Note that training is highly stable, and all hyperparameter settings yield similar results.

Our training is implemented using the \library{PyTorch} library \cite{Ansel_PyTorch_2_Faster_2024}, the \library{Transformers} library \cite{Wolf_Transformers_State-of-the-Art_Natural_2020}, the \library{Datasets} library \cite{Lhoest_Datasets_A_Community_2021}, and the \library{DeepSpeed} library \cite{10.1145/3394486.3406703}.  
Experiments are conducted on servers equipped with two Intel Xeon E5-2698 v4 CPUs and eight NVIDIA V100 GPUs, each with 32GB of memory.  
Training KPR takes approximately eight hours.

\section{Results Based on Top 100 Accuracy}
\label{subsec:full-results}
Table~\ref{table:results_main_full} presents the top 100 accuracies of KPR and baseline models.  
Furthermore, Table~\ref{table:results_sota_full} reports the top 100 accuracies of KPR based on \hfmodel{bge-base} and recent off-the-shelf retrievers.

\section{Additional Experiments on MS MARCO}
\label{subsec:ms-marco}
We conduct additional experiments on the MS MARCO dataset to assess the effectiveness of KPR on a widely used retrieval benchmark.  
The experimental setup follows \citet{xiao-etal-2022-retromae}.  
We train the model on the public training set, using hard negatives mined with BM25, and employ the official corpus of 8.8 million passages as the retrieval target.  
Evaluation is conducted on the development set.  
We report Mean Reciprocal Rank (MRR@10 and MRR@100) and recall (R@10, R@100, and R@1000).

We use the hyperparameters provided in the GitHub repository of \citet{xiao-etal-2022-retromae}.\footnote{\url{https://github.com/staoxiao/RetroMAE/tree/master/examples/retriever/msmarco}}  
For each query, we select one positive and 15 negative passages to form a mini-batch of 16 queries.  
Negative passages are sampled from the top 200 ranked by BM25.  
We apply in-batch negatives during training.  
The model is trained for 4 epochs with a learning rate of 2e-5 using the Adam optimizer.  
For RetroMAE, we use the variant trained on the MS MARCO corpus\footnote{\url{https://huggingface.co/Shitao/RetroMAE_MSMARCO}}, following \citet{xiao-etal-2022-retromae}.

As shown in Table~\ref{table:results_msmarco}, KPR consistently outperforms DPR built on both BERT and RetroMAE across all metrics.  
These results clearly demonstrate the effectiveness of KPR on this benchmark.

\begin{figure*}[htb]
  \centering
  \fbox{%
    \parbox{1.0\textwidth}{
      \begingroup
       \setlist[itemize]{leftmargin=1.5em, nosep}

      \textbf{Question \#1: } in which month does the average rainfall of \hlink{new york} exceed 86 mm?
      \begin{itemize}
          \item[\cmark] New York City (\textbf{0.031})
          \item[\xmark] New York (state) (0.030)
      \end{itemize}

      \itemseprule

      \textbf{Question \#2: } what is the number of tv directors that are \hlink{jewish}?
      \begin{itemize}
          \item[\cmark] Jews (\textbf{0.041})
          \item[\xmark] Judaism (0.030)
      \end{itemize}

      \itemseprule

      \textbf{Question \#3: } how does one drink \hlink{margarita}?
      \begin{itemize}
          \item[\cmark] Margarita (\textbf{0.099})
          \item[\xmark] Margarita Island (0.061)
      \end{itemize}

      \itemseprule
      
      \textbf{Question \#4: } the largest unit of area in the \hlink{si} system is called as?
      \begin{itemize}
          \item[\cmark] International System of Units (\textbf{0.098})
          \item[\xmark] Silicon  (0.077)
      \end{itemize}
      
      \itemseprule
      \textbf{Question \#5: } \hlink{thatcher} was the inspiration for which for your eyes only character?
      \begin{itemize}
          \item[\cmark] Margaret Thatcher (\textbf{0.127})
          \item[\xmark] Thatcher, Arizona (0.071)
      \end{itemize}

      \itemseprule

      \textbf{Question \#6: } what movies were edited by \hlink{spielberg}?
      \begin{itemize}
          \item[\cmark] Steven Spielberg (\textbf{0.161})
          \item[\xmark] Spielberg, Styria (0.091)
      \end{itemize}

      \itemseprule

      \textbf{Question \#7: } who made the sensor of \hlink{d300}?
      \begin{itemize}
          \item[\cmark] Nikon D300 (\textbf{0.121})
          \item[\xmark] State road D.300 (Turkey) (0.088)
      \end{itemize}

      \itemseprule

      \textbf{Question \#8: } which position was \hlink{glen johnson} playing in 2010 world cup?
      \begin{itemize}
          \item[\cmark] Glen Johnson (\textbf{0.122})
          \item[\xmark] Glen Johnson (boxer) (0.120)
      \end{itemize}
      
      \itemseprule

      \textbf{Question \#9: } \hlink{michael tyson} uses which stance?
      \begin{itemize}
          \item[\cmark] Mike Tyson (0.107)
          \item[\xmark] Michael Tyson (antiquary) (\textbf{0.110})
      \end{itemize}
      \itemseprule

      \textbf{Question \#10: } \hlink{o} was discovered by how many people?
      \begin{itemize}
          \item[\cmark] Oxygen (0.099)
          \item[\xmark] Big O notation (\textbf{0.142})
      \end{itemize}
      \endgroup
    }
  }
  \caption{Qualitative analysis of KPR’s attention mechanism on the GraphQuestions dataset.
Entity names are underlined; correct and incorrect referent entities are marked with \cmark and \xmark, respectively.
The referent entity receiving the higher attention weight is shown in bold.
}
  \label{fig:attention-analysis}
\end{figure*}

\section{Qualitative Analysis of KPR's Attention Mechanism}
\label{subsec:analysis-attention}
To examine the behavior of KPR's attention mechanism, we conduct an experiment using the GraphQuestions dataset \cite{su-etal-2016-generating,sorokin-gurevych-2018-mixing}, which contains questions annotated with entity names linked to Wikidata entities that can be resolved to Wikipedia entities.  
We randomly select 500 questions that include entity names associated with more than two possible referent entities in the dictionary used by our entity linker.  
Each question and its entities are input to KPR, and we manually inspect the attention weights (i.e., the normalized outputs of the sigmoid activation) assigned to each entity.
 
We present 10 example questions with their entities and corresponding attention weights in Figure~\ref{fig:attention-analysis}.  
This analysis reveals two general trends:  
(1) KPR tends to assign lower attention weights to common entities, such as widely known geographic names, racial group names, and religious terms (Questions~\#1 and~\#2).  
We hypothesize that this occurs because such entities are already well-represented in the underlying BERT model, and KPR does not require additional knowledge to handle them.  
(2) KPR generally assigns higher attention weights to correct entities than to incorrect ones (Questions~\#1--\#8), suggesting that it implicitly learns to disambiguate entities.  
However, it occasionally assigns higher or comparable weights to incorrect entities, particularly when those entities are rarer than the correct ones (Questions~\#8--\#10).  
Moreover, in some cases where the correct and incorrect entities are semantically similar (e.g., Questions~\#1 and~\#2), both may still contribute positively to retrieval performance.

These findings partly explain why KPR appears robust to noise (see Section~\ref{subsec:results}), even though it sometimes assigns non-negligible weights to irrelevant entities.  
Another possible explanation is that, as suggested by a recent study~\cite{elhage2022toy}, LMs can represent numerous features simultaneously in an almost orthogonal manner, which may allow the incorporation of noisy entities without significantly affecting similarity scores.
\end{document}